\documentclass[sigconf]{acmart}

\usepackage{booktabs} 
\usepackage{algorithm}
\usepackage{algorithmicx}
\usepackage{algpseudocode}

\setcopyright{rightsretained}

\copyrightyear{2017} 
\acmYear{2017} 
\setcopyright{acmcopyright}
\acmConference{WI '17}{August 23-26, 2017}{Leipzig, Germany}\acmPrice{15.00}\acmDOI{10.1145/3106426.3106437}
\acmISBN{978-1-4503-4951-2/17/08}

\begin{document}
\title{Navigation Objects Extraction for Better Content Structure Understanding}
\author{Kui Zhao}
\affiliation{
  \institution{College of Computer Science, Zhejiang University}
  \city{Hangzhou} 
  \country{China}
}
\email{zhaokui@zju.edu.cn}

\author{Bangpeng Li}
\affiliation{
  \institution{College of Computer Science, Zhejiang University}
  \city{Hangzhou} 
  \country{China}
}
\email{bondlee@zju.edu.cn}

\author{Zilun Peng}
\affiliation{
  \institution{Department of Computer Science, University of British Columbia}
  \city{Vancouver} 
  \country{Canada}
}
\email{zilunpeng@gmail.com}

\author{Jiajun Bu}
\affiliation{
  \institution{Alibaba-Zhejiang University Joint Institute of Frontier Technologies, College of Computer Science, Zhejiang University}
  \city{Hangzhou} 
  \country{China}
}
\email{bjj@zju.edu.cn}

\author{Can Wang}
\affiliation{
  \institution{Alibaba-Zhejiang University Joint Institute of Frontier Technologies, College of Computer Science, Zhejiang University}
  \city{Hangzhou} 
  \country{China}
} 
\email{wcan@zju.edu.cn}

\begin{abstract}
Existing works for extracting navigation objects from webpages focus on navigation menus, so as to reveal the information architecture of the site. However, web 2.0 sites such as social networks, e-commerce portals etc. are making the understanding of the content structure in a web site increasingly difficult. Dynamic and personalized elements such as top stories, recommended list in a webpage are vital to the understanding of the dynamic nature of web 2.0 sites. To better understand the content structure in web 2.0 sites, in this paper we propose a new extraction method for navigation objects in a webpage.  Our method will extract not only the static navigation menus, but also the dynamic and personalized page-specific navigation lists. Since the navigation objects in a webpage naturally come in \textit{blocks}, we first cluster hyperlinks into different blocks by exploiting spatial locations of hyperlinks, the hierarchical structure of the DOM-tree and the hyperlink density. Then we identify navigation objects from those blocks using the SVM classifier with novel features such as anchor text lengths etc. Experiments on real-world data sets with webpages from various domains and styles verified the effectiveness of our method.
\end{abstract}

%
%



\maketitle

\section{Introduction}
The explosive growth of the World Wide Web generates tremendous amount of web data 
and consequently web data mining has become an important technique for discovering useful information and knowledge. 
Among many popular topics in web data mining, extracting information 
architecture or content structures for a web site has attracted many research attention in recent years. 
Existing works mainly extract navigation menus from webpages to 
reveal the content structure of the site \cite{keller2012menuminer}. 
Many applications can be derived from the extracted content structure, including generating site map to improve information accessibility for disabled users, or providing content hierarchy in search results \cite{keller2013search} etc. 

\begin{figure}[ht!]
\centering
\includegraphics[width=4cm]{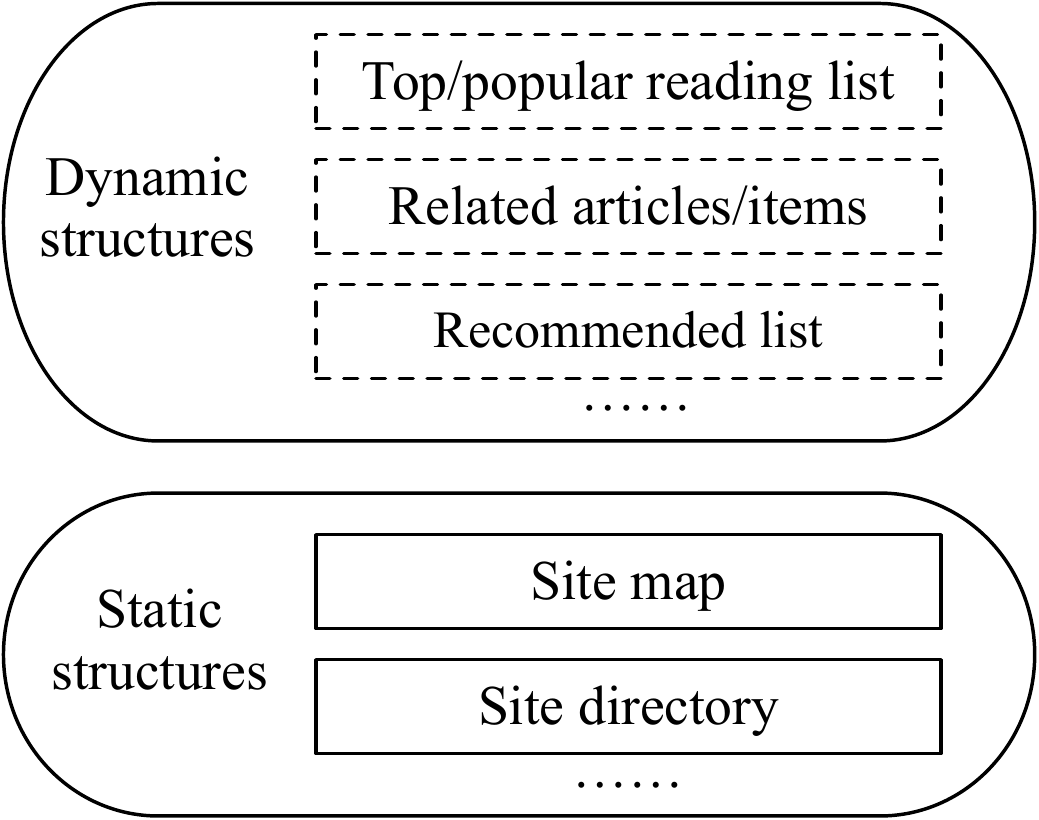}
\caption{\label{fig:info-arch}
Information architecture of a web 2.0 site.
}
\end{figure}

However, the increasing number of web 2.0 sites such as social networks, e-commerce portals etc. are 
turning the web from a static information repository into a dynamic platform for information sharing and interactions. 
As shown in Figure \ref{fig:info-arch}, the information architecture on these sites are characterized 
not only by the traditional static directory structure of the site, 
but also by the dynamic elements such as  the \textit{top reading list}, \textit{recommended items} etc. 
In fact, the dynamic nature of web 2.0 sites are better captured by these dynamic and personalized elements. 
But their importance are neglected in existing works of web structure extraction, 
which mainly focus on extracting static web site structures such as the navigation menus\cite{keller2012menuminer}, headings\cite{manabe2015extracting} etc. 

In this paper, we propose a new extraction method for navigation objects 
in a webpage to capture both the static directory structures and the dynamic content structures in a web site. 
It is a non-trivial task mainly because of the great diversities in webpage structures. Webpages come with 
various layouts, thus navigation objects in different webpages varies greatly in their presentation. 
Moreover, many navigation elements in a webpage nowadays  are generated dynamically, or customized for specific users. 

To overcome these difficulties, we attempt to develop a page-dependent extractor for navigation objects in a webpage. 
Our method is based on following observations for navigation objects, in a typical webpage: 
1) the navigation objects are naturally grouped in different \textit{hyperlink blocks}, in which few other contents other than these hyperlinks exist; 2) the anchor text for these hyperlinks are usually short and well aligned. 
With these observations, the first step of our method is to cluster hyperlinks in a webpage into multiple blocks by exploiting 
features such as spatial locations of hyperlinks, the hierarchical structure of the DOM-tree and the hyperlink density etc. 

Then we identify navigation objects using the SVM classifier. 
Generally, the hyperlink blocks in a webpage can be divided into the following four categories:
\begin{itemize}
\item {\it Navigation Menu}. Hyperlinks provide site-level navigation. 
They stay relatively invariant and can be directly mapped to the static directory structure in a website.
\item{\it Navigation List}. Hyperlinks provide page-dependent navigation and capture the dynamic and personalized content structures, 
such as recommended list etc.  
\item{\it Content Hyperlink}. Hyperlinks appears in the main content. 
\item{\it Others}. Hyperlinks include Ads, copyright information etc. 

\end{itemize}
Obviously, we intend to extract {\it Navigation Menu} and {\it Navigation List} in a webpage. 
The SVM classifier is trained with some well defined features, 
such as the number of hyperlinks, the mean and the variance of anchor text lengths etc. 
Experimental results in multiple real-world datasets verify the effectiveness of our method. 

The rest of the paper is organized as follows. We briefly	review related works in section 2. 
We describe our method in section 3 and 4, the part of clustering hyperlinks into blocks is in section 3 
and the part of classifying hyperlink blocks is in section 4. 
Then in section 5 we show our experimental setup and results followed by discussing the results. 
Finally, we present our conclusions and plans to future research in section 6. 

\section{Related Work}
Our work is related to areas of web structure mining and web information extraction. 

\textbf{Web structure mining}. 
Web structure mining aims to study the hyperlink structure of the web.  
Some early works studied the structure of the web at large \cite{broder2000graph}\cite{jon2001the} 
and uncover the major connected components of the web. Others analyzed the generally properties related 
with the web graph, such as its diameter \cite{albert1999internet}, 
size and accessibility of information on the web \cite{lawrence1999accessibility} etc.  
PageRank \cite{page1999pagerank} exploits the linkage information to learn the 
importance of webpages and becomes widely used in modern search engines. 

Recent works on web structure mining focused more on the local structures of the web graph.
Ravi et al. \cite{kumar2006hierarchical} used the hierarchical structure of URLs 
to generate hierarchical web site segmentation. 
Though the hierarchical structure of URLs was also used in many other works, such as \cite{yang2009web}, 
the hierarchical structure of URLs does not reflect the web site organization accurately. 
Eduarda Mendes et al. \cite{rodrigues2006link} noticed that and thought navigation objects could 
reflect the web site structure better. 
They applied frequent item-set algorithms on the 
outgoing hyperlinks of webpages to detect repeated navigation menus and then used them to represent web sites. 
Keller et al. \cite{keller2012menuminer} also tried to use navigation menus to 
reveal the information architecture of web sites, but they extracted navigation menus in a very different way. 
They extracted navigation menus by analyzing maximal cliques on the web graph. 
Some works do not extract navigation objects directly, but they take into account the 
structural information navigation objects provide. For instance, 
when Cindy Xide et al. \cite{lin2010hierarchical} clustered webpages, 
they considered {\it parallel links} which are siblings in the DOM-tree of a webpage 
and usually in the same navigation objects. 
However, these works only focus on the static structure of a web site represented by navigation menus etc. 
and neglect the dynamic structure represented by personalized page-specific navigation lists. 
These navigation elements is vital to understand the dynamic nature of web 2.0 sites.

\textbf{Web information extraction}. 
Information extraction from webpages has many applications. 
Most of the existing works focus on main content extraction from webpages and the early work about that can 
be traced back to Rahman et al. \cite{rahman2001content}. 
They segment the webpages into zones based on its HTML structure and extract important 
contents by analyzing zone attributes. 

Among many different categories of extraction methods, template-based ones are popular because they are highly accurate and easy to implement. 
They extracted content from pages with a common template by looking for special HTML cues using regular expressions. 
A different category of template-based methods used {\it template detection} algorithms 
\cite{bar2002template}\cite{lin2002discovering}\cite{yi2003eliminating}\cite{chen2006template}, 
in which webpages with the same template are collected and used to learn common structures. 
The major problem with template-based extractors is that different extractors must be developed for different templates. 
What's more, once the template updates, as frequently happens in many web site, the extractor will be invalidated. 

To overcome the limitations of template-based methods, many researchers attempted to extract 
content from webpages in a template-independent way. 
Cai et al. \cite{cai2003vips} proposed a vision-based webpage segmentation algorithm 
named VIPS to divide a webpage into several blocks by its visual presentation. 
Zheng et al. \cite{zheng2007template} presented a template-independent news extraction method based on visual consistency. 
Wang et al. \cite{wang2009can} exploited more features about the relation between the news title and body by firstly extracting 
the title block and then extracting the body block.  Shanchan et al. \cite{wu2015automatic} trained a machine learning model with multiple features generated by utilizing DOM-tree node properties and extracted content using this model.
Although these methods extract webpage content in a template-independent way, they still have to rely on some particular HTML cues (e.g., $<table>$, $<td>$, color and font etc) in their extraction, and thus are more easily affected by the underlying web development technologies.
Two recent works, CETR \cite{weninger2010cetr} and CETD \cite{sun2011dom} address this issue by identifying regions with 
high text density, i.e., regions including many words and few tags are more likely to be main content. 

As can be seen, most existing works of information extraction from webpages 
focus on main content extraction and they can not be applied to extracting navigation objects directly. 
Even the template-based methods cannot be used directly to extract navigation objects because 
navigation lists in webpages are usually generated dynamically and page-dependent.
\section{Clustering hyperlinks}
\begin{figure}[ht!]
\centering
\includegraphics[width=4.5cm]{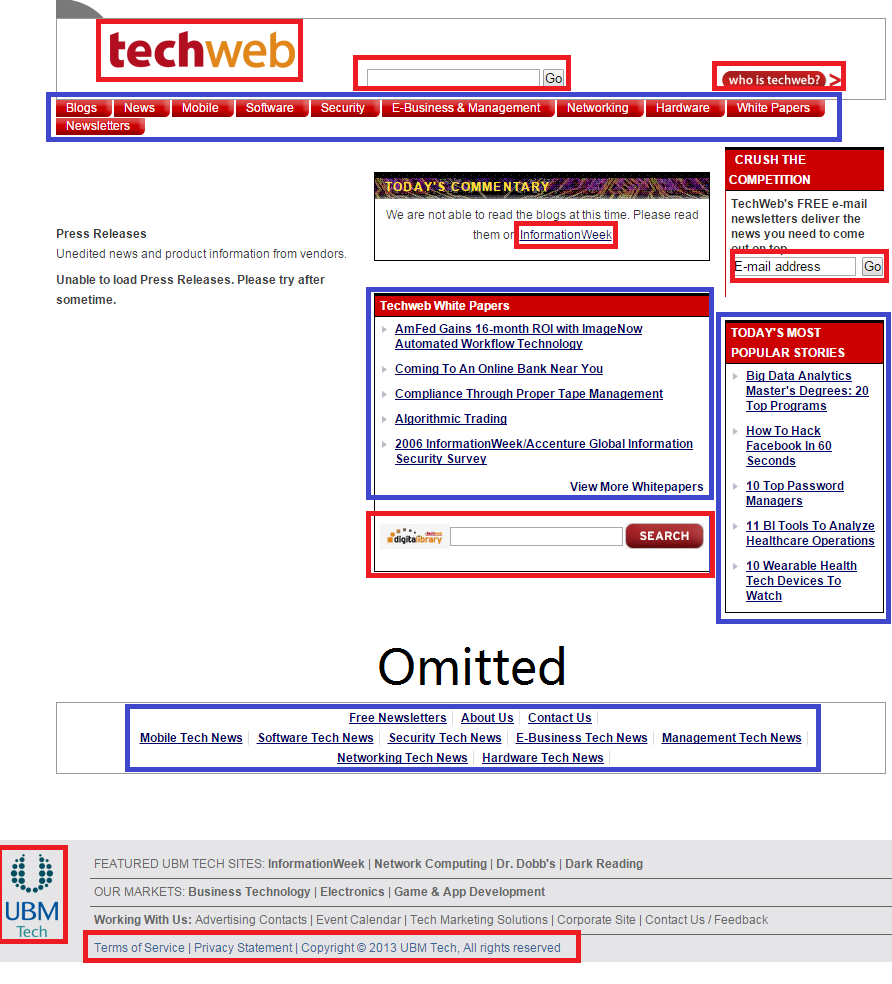}
\caption{\label{fig:techweb}
Techweb's home page. Blue boxes mark the navigation blocks and red boxes mark the non-navigation blocks.
}
\end{figure}

Our work is motivated by the observation that the navigation objects are naturally grouped in different hyperlink blocks according to their purposes. To better illustrate our idea, we use a typical webpage, the home page of  Techweb\footnote{http://www.techweb.com} as an example. As shown in Figure \ref{fig:techweb}, 
the hyperlinks in the webpage are obviously grouped in different blocks with their different visual presentation features. 

\subsection{DOM-tree}
\label{sec:dom}
\begin{figure}[ht!]
\centering
\includegraphics[width=4cm]{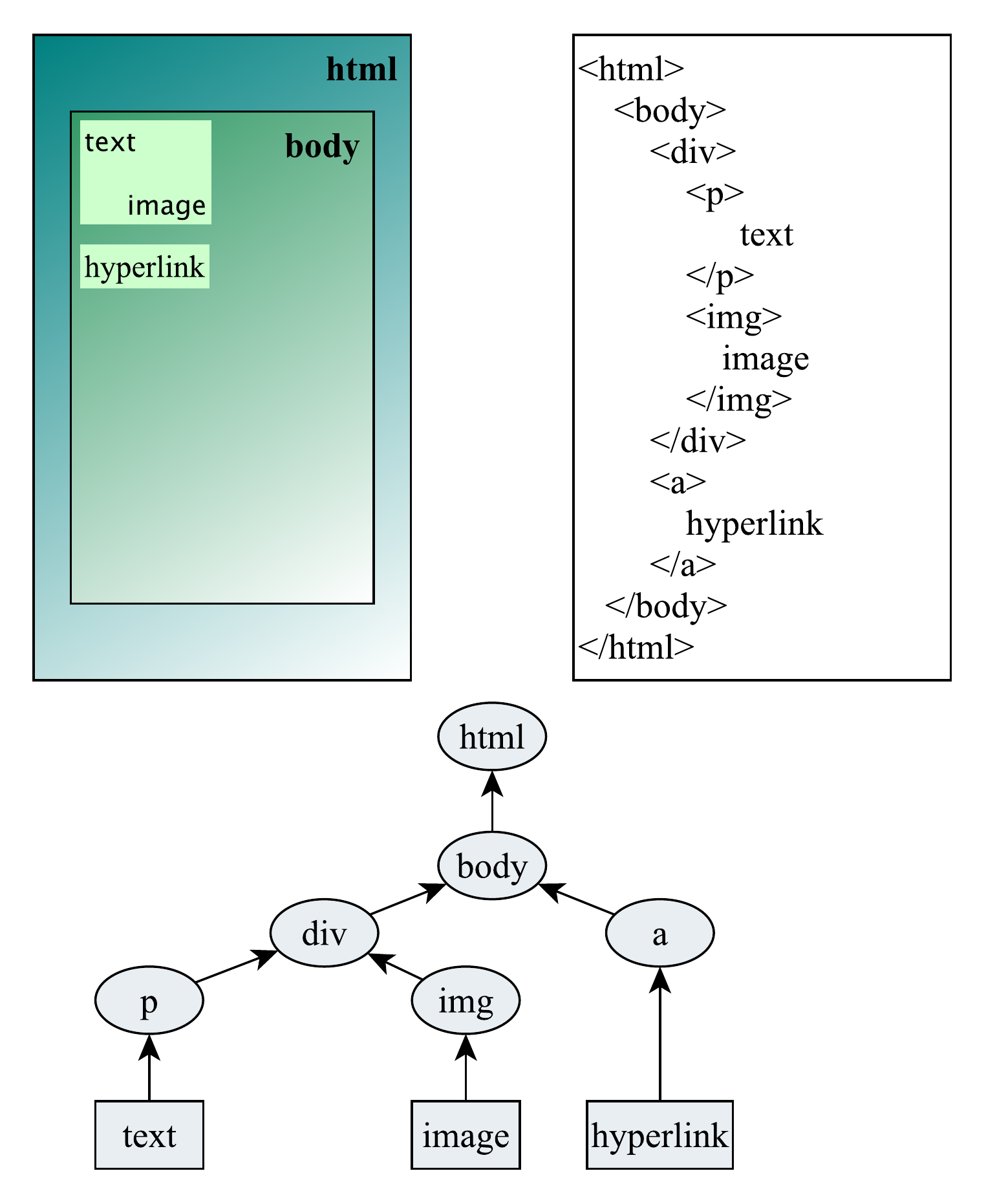}
\caption{\label{fig:dom}
An example of DOM-tree.
}
\end{figure}

Before clustering hyperlinks in a webpage into blocks, we parse the webpage into a DOM-tree. 
Each webpage corresponds to a DOM-tree where detailed text, images and hyperlinks etc. are leaf nodes. 
An example of the DOM-tree is shown in Figure \ref{fig:dom}. 
The DOM-tree at the bottom of Figure \ref{fig:dom} is derived from the HTML code 
at the top right, whose webpage layout is at the top left. 

The DOM-tree is a hierarchical structure and it has three useful properties as follows. 
First, the relation between child node and parent node reflects their relation in 
the webpage layout, e.g., in Figure \ref{fig:dom} the node $<p>$ and $<img>$ are child nodes 
of node $<div>$ reflects that {\it text} and {\it image} are included in the block corresponding to $<div>$ in the webpage layout. 
Second, the relative positions of sibling nodes are preserved when they are displaying in the webpage. 
More specifically, if node $a$ and node $b$ are sibling nodes and $a$ is at the left side of $b$ on the DOM-tree, 
the displaying element corresponding to $a$ must stay at the left side or the top of 
the displaying element corresponding to $b$ in the webpage layout. 
Third, hyperlinks in the same block must have the same ancestor, 
which is the root node of the smallest sub-tree including that block. 
The above three properties are very useful when we cluster hyperlinks into blocks on the DOM-tree of a webpage.
\subsection{DOM-tree Distance}
The central problem in clustering hyperlinks is to define a reasonable distance between them that well conforms to their visual presentation. The most intuitive choice is the Euclid distance between their locations on the webpage as rendered by browsers. However, obtaining these locations requires expensive computation cost. Moreover, locations for many hyperlinks can not be obtained without user interactions, e.g., in multilevel menus, the displaying locations of hyperlinks in the second or third level menus are only available after clicking their parent menus. 

To address this issue, we analyze the structure of the HTML code and use the \textit{DOM-tree distance} 
to approximate the distance between two hyperlinks. We first traverse the DOM-tree of a given webpage 
with depth-first search order and index each node we encounter, starting from 1. 
Then we calculate the DOM-tree Distance (DD) between hyperlinks $l_1$ and $l_2$ as follow:
\begin{equation}
\text{DD}(l_1, l_2)=|\text{\bf index}(l_1)-\text{\bf index}(l_2)|,
\end{equation}
where $\text{\bf index}(l_i)$ means the index of hyperlink $l_i$. 
\begin{figure}[ht!]
\centering
\includegraphics[width=5cm]{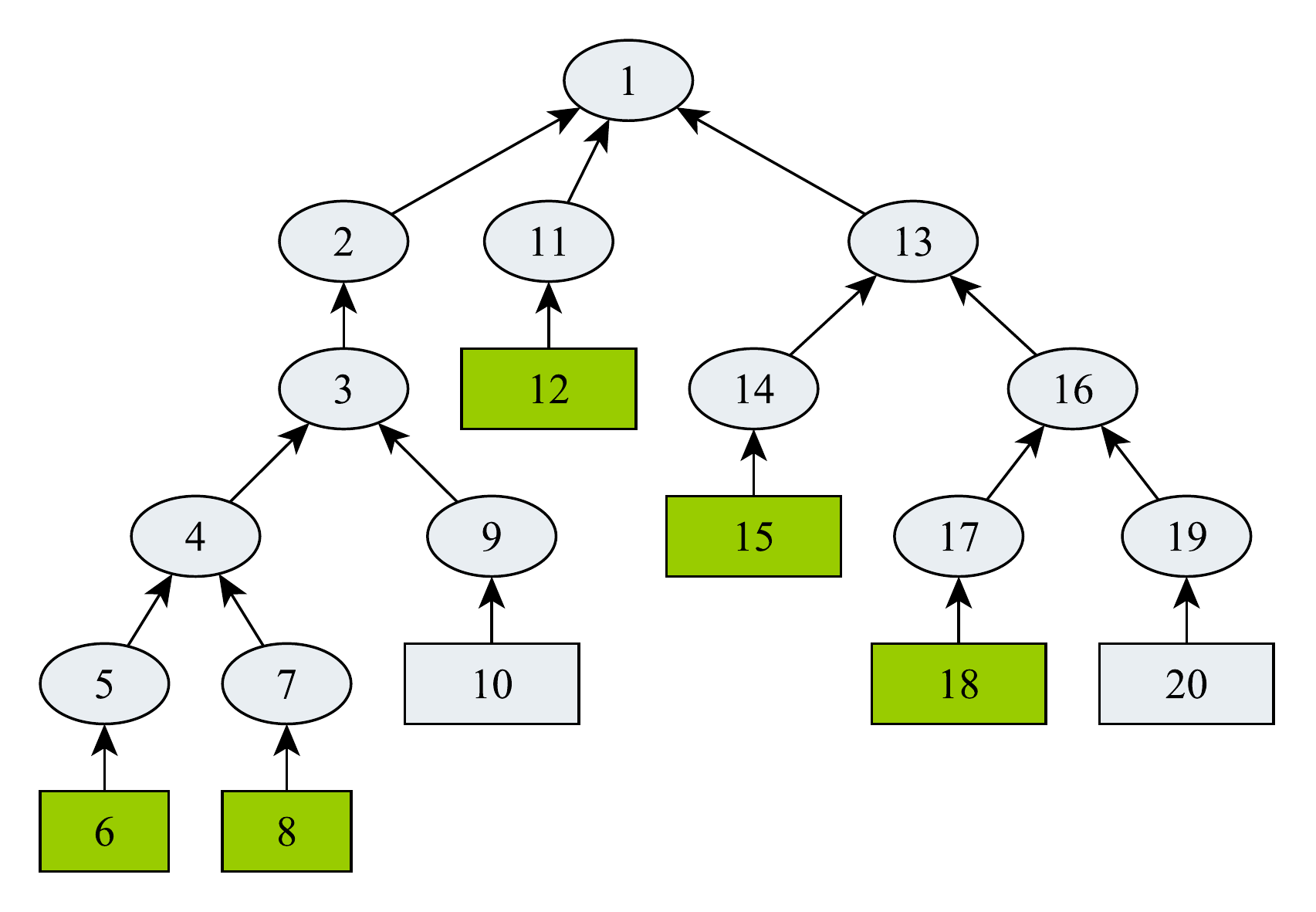}
\caption{\label{fig:dendrogram}
An example of an indexed DOM-tree, where leaf nodes in green color are hyperlink nodes.}
\end{figure}
For two given hyperlink blocks $B_1$ and $B_2$, we define the gap between them as 
the minimum distance between hyperlinks in $B_1$ and $B_2$:
\begin{equation}
\text{gap}(B_1, B_2)=\min\limits_{i,j}\text{DD}(l_i,l_j),
\end{equation}
where $l_i\in B_1$, $l_j\in B_2$. 
We can use the internal node to represent a hyperlink block, which includes all hyperlink nodes in the corresponding sub-tree. 
In Figure \ref {fig:dendrogram}, the node indexed with {\it 2} can represent the hyperlink block 
including hyperlinks indexed with {\it 6, 8} and the node indexed with {\it 11} 
can represent the hyperlink block including hyperlink indexed with {\it 12}.
The gap between these two hyperlink blocks is $\min\{4, 6\}=4$.

\subsection{Hyperlink Density}
Another important observation is that a hyperlink block usually includes few text except the text in hyperlinks.
We consequently define the {\it Hyperlink Density} $\text{HD}(S)$ for a given layout block $S$,
which consists of one or more sub-trees of a DOM-tree: 
\begin{equation}
\text{HD}(S)=\frac{\#\{\text{anchor text in }S\}+\epsilon}{\#\{\text{all text in }S\}+\epsilon},
\end{equation}
where $\#\{\text{anchor text in }S\}$ means the word number of the anchor text in all hyperlinks in $S$, 
$\#\{\text{all text in }S\}$ means the word number of all text in $S$ and $\epsilon$ is the 
smoothing parameter to avoid dividing zero. We set $\epsilon=10^{-10}$ in all our experiments.   

\subsection{Clustering on  DOM-tree}
In the process of clustering hyperlinks into blocks, we make good use of the hierarchical structure of the DOM-tree 
and its properties.  The complete algorithm of clustering hyperlinks on the DOM-tree is shown 
in Algorithm \ref{alg:clustering} with details. 
\begin{algorithm}[ht!]
	\caption{Clustering Hyperlinks on DOM-tree}
	\label{alg:clustering}
        	\begin{algorithmic}[1]
		\renewcommand{\algorithmicrequire}{\textbf{Input:}}
		\renewcommand{\algorithmicensure}{\textbf{Output:}}
            	\Require DOM-tree $T$, hyperlink nodes set $H$, Gap threshold $gt$, Hyperlink Density threshold $hdt$
		\Ensure Cluster set $C$
            	\renewcommand\algorithmicensure {\textbf{Initialization:} }
		\Ensure {$C={\varnothing}$};
         \Function{Cluster}{$root$}
         		\If {(leaf nodes of $root$)$\cap H$ is $\varnothing$}
			\State \Return {TRUE}
		\EndIf
         		\State $cList = []$; $jList = []$
                	\ForAll{$child$ of $root$ from left to right}
			\State $j=$ \Call{Cluster}{$child$}
			\State Append $j$ to $jList$; Append $child$ to $cList$
		\EndFor
		\State $cluster =[]$; $tcList = []$; $isOne=$TRUE
		\State $cNum \gets \text{the length of }cList$; $s=1$
		\For{$i = 1\to cNum$}
			\State Add $cList[i]$ into $tcList$
			\If{(leaf nodes of $cList[i]$)$\cap H$ is not $\varnothing$}
				\If{jList[i] is FALSE}
					\State Add $cluster$ into $C$
					\State $isOne=$FALSE; $cluster = []$; $tcList = []$
					\State Continue
				\EndIf
				\If {$cluster$ is not empty}
					\State $g=\text{gap}(cList[s], cList[i])$ 
					\State $hd=\text{HD}(tcList)$; $s=i$
					\If{$g>gt$ or $hd<hdt$}
						\State Add $cluster$ into $C$
						\State $isOne=$FALSE; $cluster = []$; $tcList = []$
					\EndIf
				\EndIf
				\State Add (leaf nodes of $cList[i]$)$\cap H$ into $cluster$
				\State Add $cList[i]$ into $tcList$
			\EndIf
		\EndFor
		\State Add $cluster$ into $C$
		\State \Return $isOne$
         	\EndFunction
	\State 
         \If {\Call{Cluster}{\text{the root node of T}} is TRUE}
     		\State Add $H$ to $C$
	\EndIf
	\end{algorithmic}
\end{algorithm}

The core of our algorithm is a recursive process. For two given hyperlink blocks $B_1$ and $B_2$, 
in which the hyperlinks have been ensured in the same block respectively. 
If these two hyperlink blocks have the same parent and are neighbors, we try to merge them. 
When the gap between hyperlink blocks $B_1$ and $B_2$ is no larger than a given threshold $gt$ and the Hyperlink Density of the 
potential hyperlink block consisting of $B_1$ and $B_2$ is no smaller than a given threshold $hdt$, we merge them into one hyperlink block. 
We only try to merge hyperlink blocks which have the same parent because hyperlinks in the same block should have the same ancestor. 
We only try to merge hyperlink blocks which are neighbors because the relative positions of sibling nodes are preserved 
when displaying in the webpage layout. 

The whole process executes from bottom to top on the whole DOM-tree and from left to right on each level of the DOM-tree. 
We have avoided a lot of  useless comparison by making good use of the hierarchical structure and properties of the DOM-tree.
 
\subsection{Threshold}
We use the gap threshold (denoted by $gt$) and the Hyperlink Density threshold (denoted by $hdt$) to control the results of clustering. 
Due to the variety of webpages, $gt$ and $hdt$ vary greatly for different webpages. 
So we need an effective method to learn proper $gt$ and $hdt$ for each webpage. 
\subsubsection{Gap threshold}
\label{sec:gt}
As we have explained in the previous sub-section, 
we only try to merge hyperlink blocks which are neighbors. 
So the proper value of $gt$ is among the gaps between all neighbor hyperlink blocks with an additional 0. 
Though we cannot directly get the set $S_b$ of all gaps between neighbor hyperlink blocks, 
we can easily get the set $S_h$ of all distances between neighbor hyperlinks and we now prove that $S_b=S_h$. 
Firstly, each hyperlink is a hyperlink block which only contains one hyperlink, so $S_h\subset S_b$ . 
Secondly, as defined in equation (\ref{eq:gt}),  the gap between two hyperlink blocks 
is the minimum distance between hyperlinks in those two hyperlink blocks, 
which must be the distance between two neighbor hyperlinks, so $S_b\subset S_h$. 
Above all, $S_b=S_h$ is proved. 

Let $DL$ denote $S_h$ with an additional 0, the problem of calculating $gt$ becomes choosing a proper value from $DL$:

(1) The $gt$ should not be too large to avoid clustering all hyperlinks into very few big blocks;

(2) The $gt$ should not be too small to avoid clustering all hyperlinks into too many small blocks. 

We choose the following $i$-th value in $DL$ as $gt$ after sorting $DL$ in decreasing order:
\begin{equation}
\label{eq:gt}
\arg_{i}\min\left(\frac{DL_i}{DL_1}+\beta\frac{i}{\text{length}(DL)}\right)
\end{equation}
where the $DL_1$ is the maximum value in $DL$, the $\text{length}(DL)$ is the number of values in $DL$, 
$1\le i \le \text{length}(DL)$. They are used to normalize the value of distance and the number of potential blocks. 
$\beta$ is a tradeoff parameter and we set $\beta = 1$ in all our experiments.  

\subsubsection{Hyperlink density threshold}
\label{sec:hdt}
A hyperlink block includes few text except the text in hyperlinks. 
Intuitively, since the node with $<body>$ tag is the root node of the DOM-tree 
and it contains no less other text than each hyperlink block. 
Let $\text{HD}_B$ denote the Hyperlink Density of the whole webpage, then 
\begin{equation}
\label{eq:hdt}
hdt=\gamma \text{HD}_B
\end{equation}
perform the lower bound of Hyperlink Density of hyperlink blocks. 
$\gamma\ge0$ is a tuning parameter and we set $\gamma = 1$ in our experiments. 

\section{Classifying Hyperlink Blocks}
\label{classify}
We train a SVM classifier using RBF kernel with some well defined features to identify navigation objects. 
\subsection{Features}
\subsubsection{The number of hyperlinks}
From our observation, the navigation object usually contains many hyperlinks, while other hyperlink blocks contain less hyperlinks. 
So the number of hyperlinks is a very useful feature to distinguish navigation object from non-navigation object. 
For a given hyperlink block $B_i$, we denote the number of hyperlinks in it as $\#B_i$.

\subsubsection{Text length in hyperlinks}
The length of anchor text is also very useful. 
On one hand, anchor texts in a navigation object are usually short, 
while hyperlinks in main content usually have relatively longer texts and hyperlinks 
in Ads etc. usually contain images without any text. 
So the mean of text length in a navigation object is usually small but not zero. 
On the other hand, the text in a navigation object is usually neat and the variance of these text lengths is small. 
For a given hyperlink block $B_i$, we denote the mean and variance of the text length in its hyperlinks 
as $B_i^{tm}$ and $B_i^{tv}$ respectively. 
We apply the re-implemented Gaussian smoothing \cite{weninger2010cetr} to the text lengths of hyperlinks in a DOM-tree 
to avoid sudden changes in the text lengths. 

Above all, for a given hyperlink block $B_i$, the feature vector of $B_i$ is $[\#B_i, B_i^{tm}, B_i^{tv}]$. 
Then the SVM classifier with RBF kernel is applied to classify $B_i$ as navigation object or non-navigation object. 

\subsection{SVM Classifier}
Support Vector Machine (SVM) is a famous supervised learning model.
In order to perform non-linear classification, we use the SVM classifier with RBF kernel \cite{chapelle1999support}.

When using SVM classifiers, we need to calculate the distance between two points. 
Since the ranges of different features are significantly widely different, 
the features are normalized so that each feature contributes approximately in an equal proportion to the final distance. 
What's more, the normalization can also reduce the training time of SVM classifiers \cite{tax2000feature}.
\section{Experiment}
\label{experiment}
Experiments on real world dataset demonstrate the effectiveness of our method. 
\subsection{Date Set}
In our experiments we use data from two sources: (1) dataset from CleanEval\cite{baroni2008cleaneval}; 
(2) news site data from MSS\cite{pasternack2009extracting}. 

{\bf CleanEval:} CleanEval is a shared competitive evaluation on the topic of cleaning arbitrary webpages
\footnote{http://cleaneval.sigwac.org.uk}.  
It is a diverse dataset, only a few webpages are used from each site and the sites use various styles and structures. 
Moreover, this data set has many webpages including dynamic and page-dependent navigation elements. 

{\bf MSS:} The dataset can be retrieved from Pasternak and Roth's repository\footnote{http://cogcomp.cs.illinois.edu/Data/MSS/}. 
This data set contains 45 individual websites which are further separated into two non-overlapping sets. 1) the {\it Big 5}: 
Tribune, Freep, Ny Post, Suntimes and Techweb; 2) the Myriad 40: the webpages which 
were chosen randomly from the Yahoo! Directory. 
The {\it Big 5} includes five most popular news sites and the Myriad 40 
contains an international mix of 40 English-language sites of widely varying size and 
sophistication. 

\subsection{Performance Metrics}
\subsubsection{Clustering hyperlinks}
The results of clustering hyperlinks are identifications of several hyperlink blocks, 
and we compared them with the hand-labeled ground truth.

The first metric is the Adjusted Rand Index (ARI)  \cite{hubert1985comparing}. 
Rand Index (RI) is used to measure the agreement between the output results of clustering and the ground truth\cite{rand1971objective}. 
ARI is a adjusted-for-chance version of the Rand Index, which equals 0 on average for random partitions and 1 
for two identical partitions. So the larger ARI value means the better performance. 

The second metric is the Adjusted Mutual Information (AMI) \cite{vinh2010information}. Mutual Information (MI) is 
a symmetric measurement for quantifying the statistical information shared between the output results of clustering 
and the ground truth \cite{cover2012elements}. 
AMI is an adjustment of the MI to account chances, it ranges from 0 to 1 and larger value indicates better performance.

\subsubsection{Classifying hyperlink blocks}

The performance of classifying hyperlink blocks is measured by standard metrics. 
Specifically, precision, recall and $\text{F}_1\text{-score}$ are calculated by comparing the output 
of our method against a hand-labeled gold standard. 

Performances on each dataset are calculated by averaging the scores of above metrics over all webpages.
Note that every hyperlink in the webpage is considered as a distinct hyperlink even if 
some hyperlinks appear multiple times in a webpage.
 
\begin{table*}[ht!]
\caption{ARI scores for each clustering algorithm on each source. Winners are in bold.}
\label{tb:ari}
\begin{center}
\begin{tabular}{c||c|c|c|c|c|c|c||c}
\hline
        & Tribune & NY Post & Suntimes & Freep & Techweb  & CleanEval   & Myriad 40 & Average\\
\hline
\hline
Agglomeration & {\bf 0.871}  & 0.975    & 0.536 & {0.846} & {0.814} &  0.725 & 0.784 & 0.793 \\
\hline
DBSCAN   & 0.816  & 0.975    & 0.536 & {0.846} & {0.814} &  0.702 & 0.784 & 0.782\\
\hline
K-Means  & 0.657  & 0.234    & 0.405 & 0.307 & {0.387} &  0.567 & 0.530 & 0.441 \\
\hline
Spectral Clustering & 0.546  & 0.262   & 0.420 & {0.319} & {0.394} &  0.404 & 0.515 & 0.409\\
\hline
\hline
CHD   & {\bf 0.871}  & {\bf 0.981}    & {0.899} & {0.845} & {0.880} &  {0.767} & {0.807} & {0.864}\\
\hline
CHD-LD   & {\bf 0.871}  & {\bf 0.981}    & {\bf 0.922} & {\bf 0.858} & {\bf 0.888} &  {\bf 0.825} & {\bf 0.828} & {\bf 0.882}\\
\hline
\end{tabular}
\end{center}
\end{table*}

\begin{table*}[ht!]
\caption{AMI scores for each clustering algorithm on each source. Winners are in bold.}
\label{tb:ami}
\begin{center}
\begin{tabular}{c||c|c|c|c|c|c|c||c}
\hline
        & Tribune & NY Post & Suntimes & Freep & Techweb  & CleanEval   & Myriad 40 & Average\\
\hline
\hline
Agglomeration & {\bf 0.856}  & 0.949    & 0.518 & {0.834} & {0.757} &  0.725 & 0.778 & 0.774 \\
\hline
DBSCAN   & 0.794  & 0.949    & 0.518 & {0.834} & {0.757} &  0.673 & 0.778 & 0.758\\
\hline
K-Means  & 0.771  & 0.417    & 0.591 & 0.608 & {0.603} &  0.662 & 0.643 &  0.614\\
\hline
Spectral Clustering & 0.690  & 0.439   & 0.590 & {0.627} & {0.609} &  0.508 & 0.632 & 0.585\\
\hline
\hline
CHD   & {0.855}  & {\bf 0.964}    & {0.799} & {0.836} & {0.828} &  {0.743} & {0.804} & {0.833}\\
\hline
CHD-LD   & {0.855}  & {\bf 0.964}    & {\bf 0.812} & {\bf 0.846} & {\bf 0.838} &  {\bf 0.802} & {\bf 0.831} & {\bf 0.850}\\
\hline
\end{tabular}
\end{center}
\end{table*}
\subsection{Implementation Details}
All programs were implemented in Python language with the help of scikit-learn \cite{pedregosa2011scikit}. 
After parsing the HTML file of a webpage into a DOM-tree, we treated all elements with the tag $<a>$ as hyperlinks, 
including some buttons and drop-down lists. We kept everything in a webpage without any preprocess, 
in order to show that our method can handle most noise in the webpage.
\subsubsection{Clustering hyperlinks}
In order to properly evaluate the performance of our method on clustering hyperlinks, we compared our method's
performance with several common clustering algorithms, 
including Agglomeration, DBSCAN, K-Means and Spectral Clustering \cite{rokach2010survey}. 
All algorithms use equation (1) to measure the distance between two hyperlinks. 
The Agglomeration initializes every hyperlink to a singleton cluster at the beginning. 
At each of the $N-1$ steps, the two closest clusters are merged into one singleton cluster. 
We implement this algorithm by ourselves and use {\it single-linkage} to measure the intergroup dissimilarity 
and use $gt$ as the threshold to jump out of its iteration. 
The DBSCAN algorithm regards clusters as areas with high density separated by 
areas with low density. We use the implementation in scikit-learn by setting $\text{\bf eps} = gt, \text{\bf min\_samples}=1$,
where $\text{\bf eps}$ means the maximum distance between two samples for them to be considered as 
in the same neighborhood and $\text{\bf min\_samples}$ is the minimum number of samples in 
a neighborhood for a point to be considered as a core point. 
The K-Means algorithm clusters data by trying to split samples into $K$ groups. 
We use the implementation in scikit-learn by setting parameter $K$ with the number of blocks in the ground truth. 
For Spectral Clustering ,we use the implementation in scikit-learn by setting parameter $K$ with the number 
of blocks in the ground truth and use the one nearest neighbor method to construct the affinity matrix for Spectral Clustering. 

There are two versions of our method. CHD is the version of clustering hyperlink on DOM-tree 
without considering Hyperlink Density by setting $\gamma = 0$ in equation (\ref{eq:hdt}) and  
CHD-HD is the version considering Hyperlink Density by setting $\gamma = 1$.
\subsubsection{Classifying hyperlink blocks}
The standard deviation is $\sigma=2$ in the re-implemented Gaussian smoothing algorithm.
We classify each hyperlink block as navigation object or non-navigation object by using SVM with RBF kernel implemented in scikit-learn. 
The parameters in this SVM classifier are set as $C=1.0 and \gamma =0.1$, where $C$ is the penalty 
parameter of the error term and $\gamma$ is the kernel coefficient for RBF. 

\subsection{Results}
For each data set, we randomly select 50\% webpages as the training set and the remaining webpages as the testing set. 
\subsubsection{Clustering hyperlinks}
Table \ref{tb:ari} and Table \ref{tb:ami} present the hyperlink clustering performance of  
different algorithms on the CleanEval, Myriad 40 and {\it Big 5} data sets in the ARI metric and AMI metric respectively. 
The {\it Big 5} has been broken down into it's individual sources. 

Comparing the average ARI values and AMI values over all data sets, 
our methods (including both CHD and CHD-HD) outperform all comparison methods. 
Actually, our methods have a better performance than most comparison methods when comparing 
ARI values and AMI values on individual data set. 
Moreover, our method is more reliable than the comparison methods since our method has a stable performances while comparison 
methods may collapse on some particular data sets. 
It is because that our method makes good use of the hierarchical structure of the DOM-tree as well as the distance information on DOM-tree. 
Finally, CHD-HD always performs better than CHD, especially on the dataset CleanEval, 
which has the greatest diversity and most dynamic and page-dependent navigation elements. 
That means besides the hierarchical structure of the DOM-tree, Hyperlink Density is also very helpful. 

Besides our method, the Agglomeration has the best performance, except the collapse of performance on Suntimes data set. 
Although it makes no use of any information from the hierarchical structure, it uses the fact that hyperlinks 
in a block are gathering together. The DBSCAN also uses this fact, so its performances are quite similar with 
the performance of Agglomeration. For instance, on the Myriad 40 and four sources in {\it Big 5}, the performance 
of DBSCAN is the same as Agglomeration. For K-Means and Spectral clustering, the performance is very poor, 
even though they have ``cheatet'' by using $K$ obtained from the ground truth. 
Actually, finding the best $K$ is very difficult. 

\begin{figure}[ht!]
\centering
\includegraphics[width=6cm]{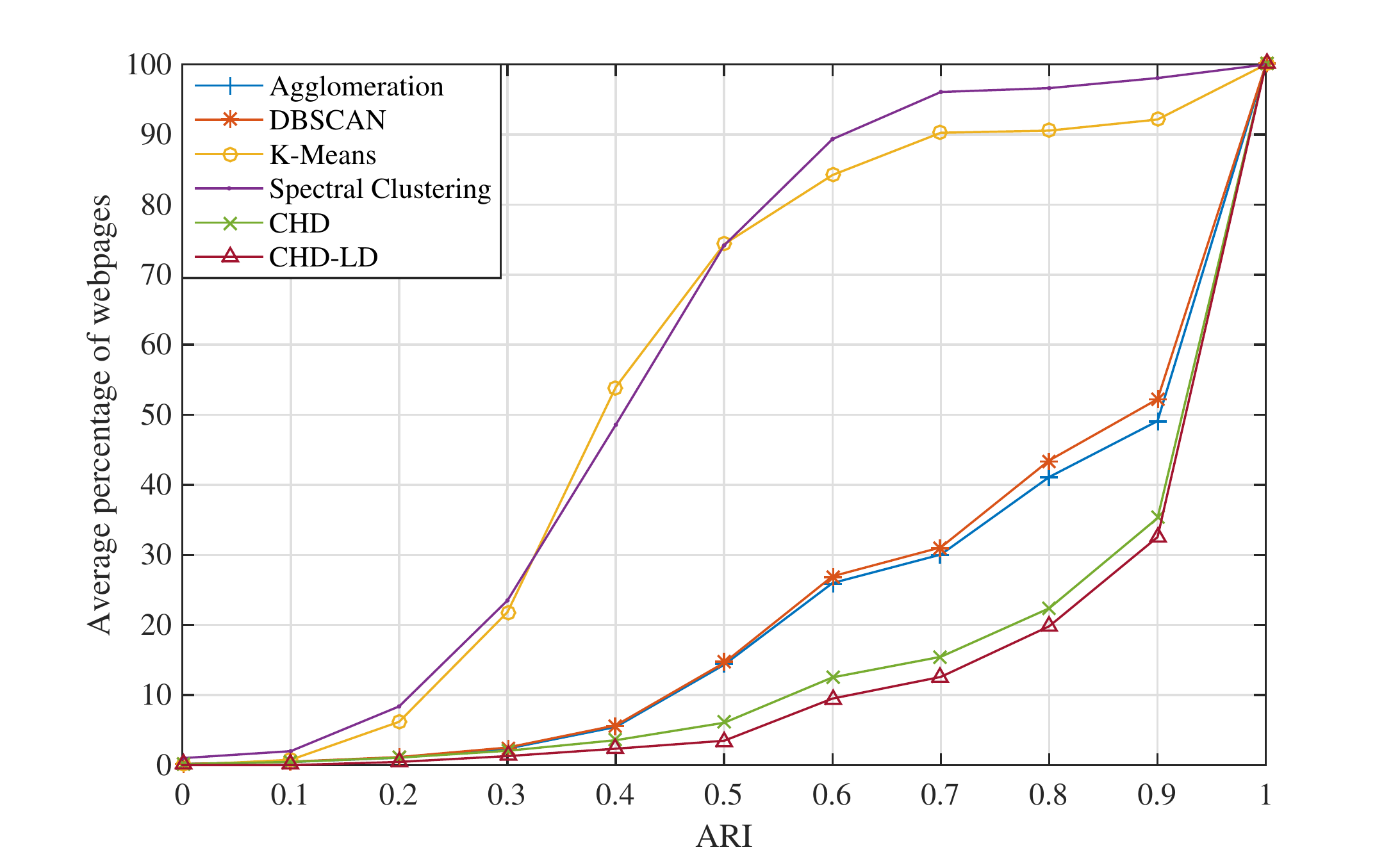}
\caption{\label{fig:b-ari}
The average percentage of webpages below ARI.}
\end{figure}

\begin{figure}[ht!]
\centering
\includegraphics[width=6cm]{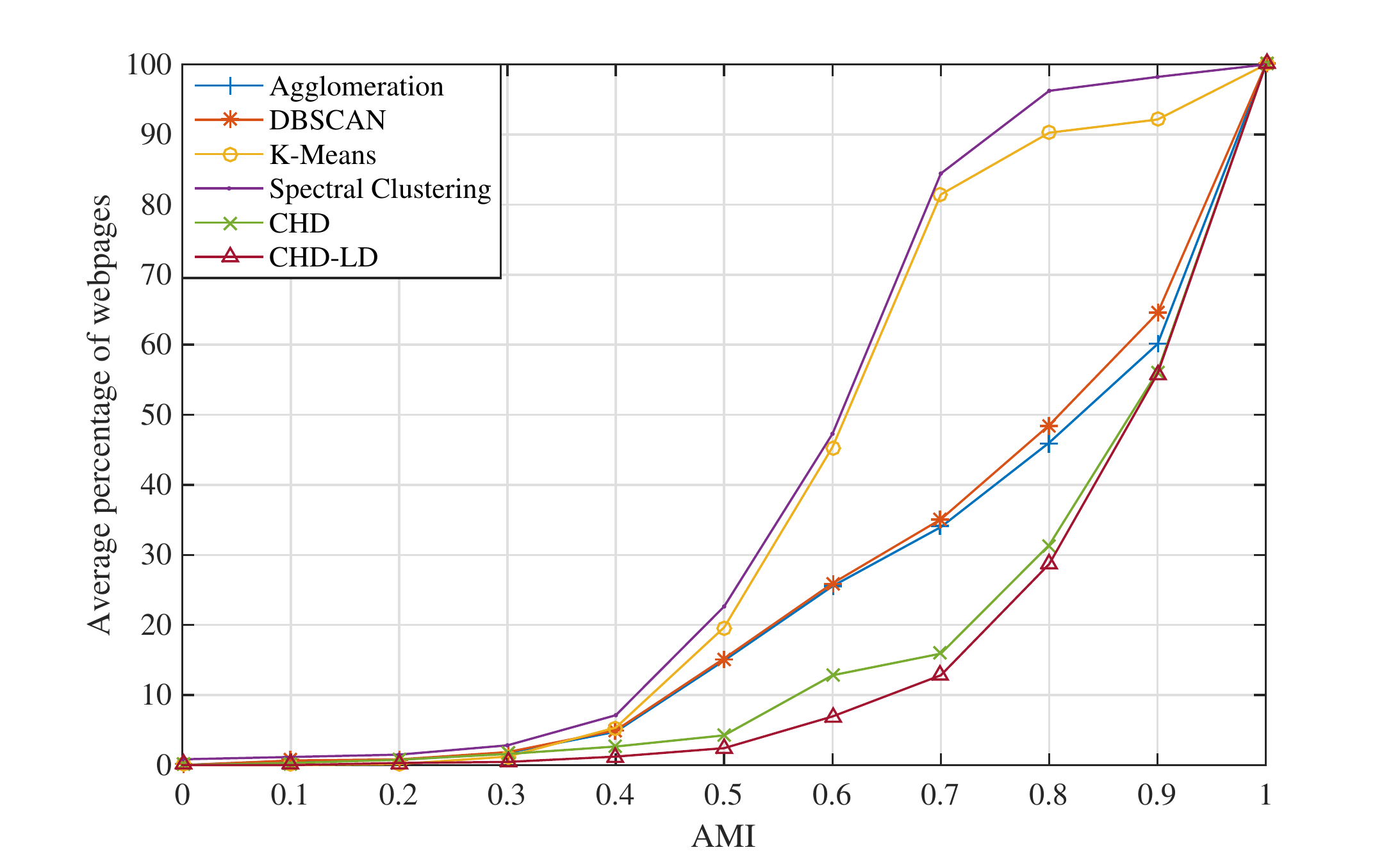}
\caption{\label{fig:b-ami}
The average percentage of webpages below AMI.}
\end{figure}

The average cumulative percentage of webpages for which the clustering performance of 
a particular method is less than a certain ARI value is plotted in Figure \ref{fig:b-ari}. 
The corresponding figure for AMI is in Figure \ref{fig:b-ami}. The slower the curve goes up from 
left to right, the better performance the corresponding method has. 
These two figures provide a more obvious illustration than Table \ref{tb:ari} and Table \ref{tb:ami}, 
in terms of the better performance that our method achieved on each ARI and AMI value relative to 
comparison methods. The majority of webpages that our method processed have a larger 
AMI or ARI value. Taking Figure \ref{fig:b-ari} as an example, 
for Agglomeration, DBSCAN, K-Means and Spectral clustering, the average percentages of ARI 
value lower than 0.6 are about 26\%, 27\%, 84\% and 89\%. At the same time, for CHD and CHD-HD, 
such percentage are only 13\% and 10\% respectively.

\subsubsection{Classifying hyperlink blocks}
Table \ref{tb:f1-chd} and Table \ref{tb:f1-chdhd} present the results of classifying hyperlink blocks. 
It clearly shows that our method performs very well, not only on datasets with webpages from 
a single site (such as Tribune and Freep etc.) but also on datasets with webpages from 
various sites (such as CleanEval and Myriad 40). The results on 
CleanEval data set are less competitive because this data set has the greatest diversity. 
Moreover, the result in which CHD-HD is used for clustering is better than 
the result under clustering using CHD. That is very reasonable because 
CHD-HD can achieve better clustering results than CHD. 
\begin{table}[ht!]
\caption{Results for extraction with clustering by CHD.}
\label{tb:f1-chd}
\begin{center}
\begin{tabular}{c||c|c|c}
\hline
 Source &  Precision  & Recall  & $\text{F}_1$-score \\
\hline
\hline
CleanEval    &  0.759 & 0.895   & 0.821 \\
\hline
Myriad 40      & 0.891  &  1.00  & 0.942\\
\hline
Tribune     & 0.870  &  0.999  & 0.930 \\
\hline
Freep    &  0.953 &  1.00  & 0.976\\
\hline 
NY Post    & 0.968  & 0.998   & 0.983\\
\hline
Suntimes   &  0.951 &  0.985  & 0.968 \\
\hline
Techweb   & 0.807  & 0.980  & 0.884\\
\hline
\end{tabular}
\end{center}
\end{table}
\begin{table}[ht!]
\caption{Results for extraction with clustering by CHD-LD.}
\label{tb:f1-chdhd}
\begin{center}
\begin{tabular}{c||c|c|c}
\hline
 Source &  Precision  & Recall  & $\text{F}_1$-score \\
\hline
\hline
CleanEval    &  0.870 & 0.801   & 0.834 \\
\hline
Myriad 40      & 0.891  &  1.00  & 0.942\\
\hline
Tribune     & 0.870  &  0.999  & 0.930 \\
\hline
Freep    &  0.955 &  0.998  & 0.976\\
\hline 
NY Post    & 0.968  & 0.998   & 0.983\\
\hline
Suntimes   &  0.934 &  0.986  & 0.959 \\
\hline
Techweb   & 0.803  & 0.989  & 0.886\\
\hline
\end{tabular}
\end{center}
\end{table}

\subsection{Discussion}
To show the generalization ability of our method, we continuously increase the percentage of hyperlinks 
in training set to be used from 1\% to 100\%, and plot the corresponding $\text{F}_{\rm 1}$-scores. 
The incremental value is 1\% when the percentage is less than 10\%, 
and the incremental value is 10\% otherwise. 
We used CHD-HD to cluster hyperlinks in this experiment. 
We can observe that even using very few hyperlinks as the training data, 
5\% hyperlinks of whole training set for an example, the performance of our method is very impressive. 
This means our method has a strong generalization ability because it needs very few training data to perform very well.  
That brings great practicability to our method. 

\begin{figure}[ht!]
\centering
\includegraphics[width=6cm]{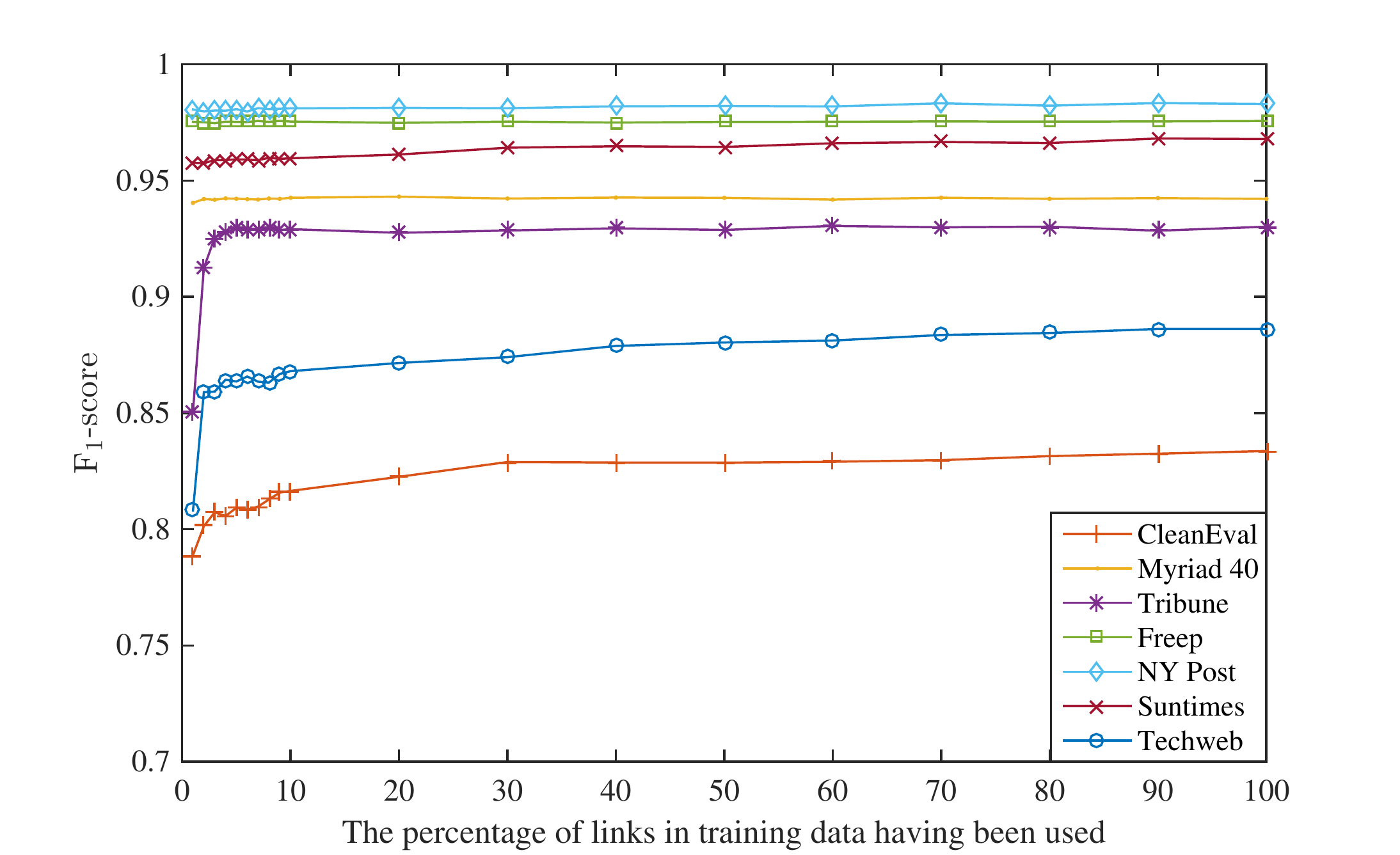}
\caption{\label{fig:f1}
The change of ${\rm F}_1$-score over the percentage of hyperlinks used as training data.}
\end{figure}

\section{Conclusions}
\label{conclusion}
In this paper we propose a new extracting method for navigation objects in a webpage 
to capture both the static directory structures and the dynamic content structures in a website.  
Our method will extract not only the static navigation menus, 
but also the dynamic and personalized page-specific navigation lists, 
including top stories and recommended list etc.
Based on the observation that hyperlinks in a webpage are naturally arranged in different blocks, 
we use a two-step process to extract navigation objects in a webpage by first clustering hyperlinks in a webpage into multiple blocks and then identify navigation object blocks from the clustering results using the SVM classifier. 
The effectiveness of our method is verified with experiments on real-world data sets. 

In addition to its effectiveness, the greatest strengths of our method are 
the simplicity of its implementation and its great practicability. 
Firstly, it has a very strong ability of generalization and needs very few training data to perform well, 
which gives it great practicability. Secondly, our method only requires the HTML file of a webpage and 
does not need any preprocess to handle noises in the webpage. 
Thirdly, our method does not rely on any special HTML cues (e.g., $<table>$, $<td>$, color and font etc.), 
which brings great stabilization over time. 

There are several interesting problems to be investigated in our future work:
(1) we will consider using more features in clustering hyperlinks and classifying hyperlink blocks without injuring the simplicity of our method; 
(2) we may try to achieve similar performance without any training data, which makes the method much easier to use; 
(3) we can incorporate additional information in our method, such as the cliques in the web graph, 
to further improve the understanding of content structures in websites.

\section*{Acknowledgments}
This work is supported by Alibaba-Zhejiang University Joint Institute of Frontier Technologies,  
Zhejiang Provincial Soft Science Project (Grant no. 2015C25053), 
Zhejiang Provincial Natural Science Foundation of China (Grant no. LZ13F020001), 
National Science Foundation of China (Grant no. 61173185). 

\bibliographystyle{ACM-Reference-Format}
\bibliography{sigproc} 

\end{document}